\documentclass[runningheads]{llncs}

\usepackage[T1]{fontenc}
\usepackage{graphicx}
\usepackage{amsmath}
\usepackage{amssymb}
\usepackage{booktabs}
\usepackage{multirow}
\usepackage{xspace}
\usepackage{tikz}
\usetikzlibrary{positioning,arrows.meta,fit,backgrounds,calc,shapes.geometric}

\usepackage[breaklinks=true]{hyperref}        
\usepackage{xurl}                             
\definecolor{cbase}{RGB}{225,234,246}
\definecolor{cbaseB}{RGB}{54,96,152}
\definecolor{cwdr}{RGB}{218,240,224}
\definecolor{cwdrB}{RGB}{38,132,72}
\definecolor{cassign}{RGB}{253,236,210}
\definecolor{cassignB}{RGB}{198,120,24}
\definecolor{closs}{RGB}{235,224,246}
\definecolor{clossB}{RGB}{120,72,158}

\newcommand{\pf}{\textit{P.\,falciparum}\xspace}
\newcommand{\pmm}{\textit{P.\,malariae}\xspace}
\newcommand{\po}{\textit{P.\,ovale}\xspace}
\newcommand{\pv}{\textit{P.\,vivax}\xspace}

\begin{document}

\title{CoSWA-YOLOv12: Scale-Invariant Tiny Object Detection and Segmentation of Malaria Parasites}

\titlerunning{CoSWA-YOLOv12: Scale-Invariant Malaria Parasite Detection}

\author{Ahmed Tahiru Issah\inst{1} \and
Carine Mukamakuza\inst{1}\thanks{Corresponding author.}}
\authorrunning{A.T. Issah et al.}
%
\institute{Carnegie Mellon University, Kigali, Rwanda\\
\email{aissah@alumni.cmu.edu, cmukamak@andrew.cmu.edu}
}

\maketitle

\begin{abstract}
Automated microscopy could widen access to malaria diagnosis in low-resource settings, but the deadliest species, \pf, presents in its early ring stage as an object only a few tens of pixels wide. Such tiny targets are systematically under-detected: overlap-based label assignment starves them of positive samples, and overlap-based box regression gives weak gradients at their scale. The Normalized Gaussian Wasserstein Distance (NWD) repairs both effects, but applied uniformly across a slide that also holds objects three to four times larger it loosens their supervision and erodes their localisation, so overall accuracy can fall even as the tiny class improves. We present \textbf{CoSWA-YOLOv12}, a compact YOLOv12 instance-segmentation detector whose core Cooperative Scale-adaptive Wasserstein Assignment routes the Wasserstein treatment to an object in inverse proportion to its size, tapering back to standard assignment for larger species. Two further components support it: a wavelet detail residual, and a min-max Gaussian regression loss ($\text{M}^2$-NWD). All three additions are \emph{transfer-safe}: each reproduces the standard pretrained model exactly at initialisation, so public pretrained weights load without any loss of accuracy. On a five-class Rwandan thick-smear dataset, CoSWA-YOLOv12 raises \pf recall from 0.63 to 0.74 and mAP@50 from 0.73 to 0.81 (mask), cuts missed \pf from 38\% to 15\%, and improves strict-localisation mAP@50-95 on all five classes for both detection and segmentation, while a $2\times2$ ablation shows the scale gate and the regression loss are synergistic.

\end{abstract}

\section{Introduction and Background}
\label{sec:intro}

Malaria remains one of the heaviest and most inequitably distributed disease burdens worldwide, with the World Health Organization estimating 263 million cases and 597{,}000 deaths in 2023, more than 90\% of them in sub-Saharan Africa~\cite{who_wmr2024}. Light microscopy of Giemsa-stained blood films is still the diagnostic reference standard in endemic regions~\cite{tangpukdee_malaria_2009,poostchi_image_2018}, but reading a slide is slow, fatiguing, and dependent on the microscopist's skill, and skilled microscopists are exactly what many endemic districts lack~\cite{delahunt_metrics_2024}. Automated analysis of digitised smears therefore offers a route to expert-level screening where experts are scarce, provided the models are accurate on the cases that matter and light enough for modest hardware.

The case that carries the most weight is also the hardest to see. \pf causes the overwhelming majority of malaria deaths, yet its early ring-stage forms are minute, often under $20\times20$ pixels even at high capture resolution, and easily confused with stain debris and platelets. Modern detectors handle the larger \textit{Plasmodium} species and white blood cells comfortably but miss a substantial fraction of these rings, the worst available failure mode: a missed parasite is a missed diagnosis. Recall on the smallest, deadliest class, not mean precision across classes, is the clinically decisive quantity~\cite{delahunt_metrics_2024}.

Two mechanisms inside single-stage detectors work against tiny objects. First, \emph{label assignment} selects positive samples by measuring overlap between candidate and ground-truth boxes; for a very small box a displacement of two or three pixels collapses the Intersection over Union, so few candidates qualify and the tiny class is starved of supervision. Second, the \emph{box-regression loss} is itself overlap-based, and its gradient weakens when boxes barely intersect, the common situation for tiny targets early in training. The Normalized Gaussian Wasserstein Distance (NWD) addresses the first by modelling boxes as Gaussians whose similarity decays smoothly rather than collapsing~\cite{wang_normalized_2021}, and min-max penalties such as M2IoU sharpen the second~\cite{shandilya_m2iou_2024}, but both are applied uniformly. On a multi-species smear that mixes a 34-pixel ring with a 100-pixel gametocyte or leukocyte, the same treatment that rescues the ring loosens assignment for the large object and degrades its localisation: a scale conflict in which gains on the tiny class are paid for by losses on the rest.

Our position is that the treatment should follow the target. The central contribution is a \emph{scale-adaptive} Wasserstein assignment that applies the Gaussian similarity to an object in proportion to how small it is, tapering smoothly back to standard overlap assignment as objects grow, so tiny parasites receive the treatment they need while larger species keep the tight supervision they already had. We build this mechanism, with two supporting components, into a compact detector we call \textbf{CoSWA-YOLOv12}, after its core Cooperative Scale-adaptive Wasserstein Assignment. It extends YOLOv12 ~\cite{tian_yolov12_2025} and is what we call \emph{transfer-safe}: each added component is initialised so that, before any training, the modified network computes exactly the same function as the standard pretrained model. Publicly available pretrained weights therefore load with no loss of accuracy, and the additions can only improve the model as they train, never degrade it, which matters when locally collected data are scarce (Section~\ref{sec:impact}).

\paragraph{Contributions.}
We contribute \emph{(i)} a scale-adaptive Wasserstein label assignment that gates the tiny-object treatment by object size through a single scalar per object, delivering the NWD benefit to ring-stage \pf while removing the collateral damage uniform NWD inflicts on larger classes; \emph{(ii)} two supporting transfer-safe components, a wavelet detail residual (WDR) that preserves high-frequency structure through strided downsampling and a min-max Gaussian regression loss ($\text{M}^2$-NWD) that carries the M2IoU idea into the Wasserstein domain, each collapsing exactly to the baseline at initialisation; and \emph{(iii)} a controlled factorial study on multi-species thick-smear microscopy that reports every standard metric for detection and segmentation, shows improvements in strict-localisation mAP@50-95 on all five classes and large gains in \pf recall and mAP@50, and demonstrates that the gate and the loss act synergistically.

\section{Previous Work}
\label{sec:related}

\subsubsection{Deep learning for malaria microscopy.}
Single-stage detectors of the YOLO family are the workhorse for automated parasite analysis: they process a whole field of view, localise and classify parasites jointly, and run fast enough for point-of-care use~\cite{koirala_deep_2022,zedda_yolo-pam_2023}. Attention and parasite-specific modules raise sensitivity~\cite{zedda_deep_2025}, and instance segmentation adds pixel-level masks that separate parasites from thick-smear debris~\cite{ronneberger2015unet}. One finding recurs: early ring-stage \pf is the limiting case, the smallest target, the most readily confused with staining artefacts, and the most vulnerable to acquisition differences between laboratories~\cite{delahunt_metrics_2024}.

\subsubsection{Tiny object detection.}
The dominant response to small objects is to preserve resolution: a high-resolution detection head at stride four recovers fine structure and helps small targets~\cite{mura_yolo-tryppa_2025,koirala_deep_2022}. A complementary line targets the training signal: NWD replaces IoU with a Gaussian similarity that does not collapse for tiny boxes~\cite{wang_normalized_2021}, while M2IoU weights the box corners asymmetrically through a min-max term and converges faster than CIoU on medical detection~\cite{shandilya_m2iou_2024}. Both act on every object regardless of size; CoSWA-YOLOv12 keeps their tiny-object benefits but makes the Wasserstein treatment conditional on size, so one multi-scale model no longer trades large-object accuracy for small-object gains.

\section{Methodology}
\label{sec:method}

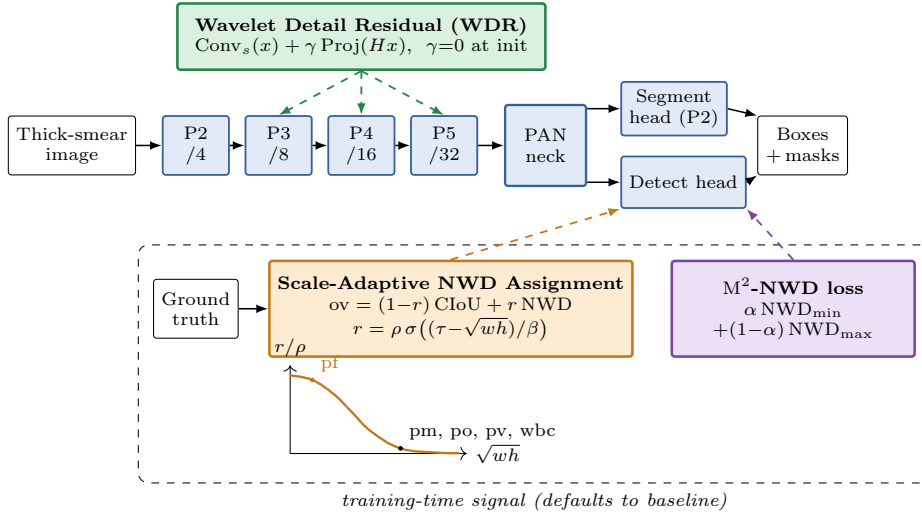
\begin{figure}[t]
\centering
\resizebox{\textwidth}{!}{%
\begin{tikzpicture}[
  font=\scriptsize,
  io/.style={draw, rounded corners=1.2pt, minimum height=8mm, minimum width=11mm, align=center, fill=white, line width=0.4pt},
  stage/.style={draw=cbaseB, line width=0.7pt, rounded corners=1.2pt, minimum height=8mm, minimum width=9mm, align=center, fill=cbase},
  neck/.style={draw=cbaseB, line width=0.9pt, rounded corners=1.2pt, minimum height=11mm, minimum width=11mm, align=center, fill=cbase},
  head/.style={draw=cbaseB, line width=0.7pt, rounded corners=1.2pt, minimum height=7mm, minimum width=14mm, align=center, fill=cbase},
  wdr/.style={draw=cwdrB, line width=1pt, rounded corners=1.2pt, align=center, fill=cwdr},
  assign/.style={draw=cassignB, line width=1pt, rounded corners=1.2pt, align=center, fill=cassign},
  loss/.style={draw=clossB, line width=1pt, rounded corners=1.2pt, align=center, fill=closs},
  ar/.style={-{Latex[length=1.6mm]}, line width=0.6pt},
  dar/.style={-{Latex[length=1.6mm]}, line width=0.7pt, dashed},
]
\node[io] (in) {Thick-smear\\image};
\node[stage, right=3.5mm of in] (s2) {P2\\{$/4$}};
\node[stage, right=2mm of s2] (s3) {P3\\{$/8$}};
\node[stage, right=2mm of s3] (s4) {P4\\{$/16$}};
\node[stage, right=2mm of s4] (s5) {P5\\{$/32$}};
\node[neck, right=3.5mm of s5] (neck) {PAN\\neck};
\node[head, right=4.5mm of neck, yshift=5mm] (seg) {Segment\\head (P2)};
\node[head, right=4.5mm of neck, yshift=-5mm] (det) {Detect head};
\node[io, right=4mm of seg, yshift=-5mm] (pred) {Boxes\\$+$\,masks};
\draw[ar] (in) -- (s2);
\draw[ar] (s2)--(s3);\draw[ar] (s3)--(s4);\draw[ar] (s4)--(s5);
\draw[ar] (s5) -- (neck);
\draw[ar] (neck.east|-seg.west) -- (seg.west);
\draw[ar] (neck.east|-det.west) -- (det.west);
\draw[ar] (seg.east) -- (pred.north west);
\draw[ar] (det.east) -- (pred.south west);
\node[wdr, above=6mm of s4, minimum width=50mm, minimum height=9mm] (wdrbox)
  {\textbf{Wavelet Detail Residual (WDR)}\\
   $\mathrm{Conv}_s(x)+\gamma\,\mathrm{Proj}(Hx)$,~~$\gamma{=}0$ at init};
\draw[dar, cwdrB] (wdrbox.south) -- (s3.north);
\draw[dar, cwdrB] (wdrbox.south) -- (s4.north);
\draw[dar, cwdrB] (wdrbox.south) -- (s5.north);
\node[io, below=14mm of s2, anchor=north] (gt) {Ground\\truth};
\node[assign, right=4mm of gt, minimum width=46mm, minimum height=13mm] (asg)
  {\textbf{Scale-Adaptive NWD Assignment}\\
   $\mathrm{ov}=(1{-}r)\,\mathrm{CIoU}+r\,\mathrm{NWD}$\\
   $r=\rho\,\sigma\big((\tau{-}\sqrt{wh})/\beta\big)$};
\node[loss, right=5mm of asg, minimum width=33mm, minimum height=13mm] (los)
  {\textbf{$\mathrm{M}^2$-NWD loss}\\
   $\alpha\,\mathrm{NWD}_{\min}$\\$+(1{-}\alpha)\,\mathrm{NWD}_{\max}$};
\draw[ar] (gt.east) -- (asg.west);
\draw[dar, cassignB] (asg.north) -- (det.south west);
\draw[dar, clossB] (los.north) -- (det.south east);
\begin{scope}[shift={($(asg.south west)+(3mm,-13mm)$)}]
  \draw[->, line width=0.4pt] (0,0) -- (2.4,0) node[right, font=\scriptsize] {$\sqrt{wh}$};
  \draw[->, line width=0.4pt] (0,0) -- (0,1.2) node[above, font=\scriptsize] {$r/\rho$};
  \draw[cassignB, line width=0.9pt] plot[smooth, tension=0.7] coordinates
    {(0,1.06)(0.3,1.0)(0.6,0.80)(0.85,0.55)(1.1,0.30)(1.5,0.07)(2.0,0.012)(2.3,0.006)};
  \fill[cassignB] (0.3,1.0) circle (0.9pt) node[above right, font=\scriptsize] {pf};
  \fill[black] (1.5,0.07) circle (0.9pt) node[above right, font=\scriptsize] {pm, po, pv, wbc};
\end{scope}
\coordinate (gcbot) at ($(asg.south)+(0,-15mm)$);
\begin{scope}[on background layer]
\node[draw, dashed, rounded corners, fit=(gt)(asg)(los)(gcbot), inner sep=2mm,
  label={[font=\scriptsize\itshape]below:training-time signal (defaults to baseline)}] {};
\end{scope}
\end{tikzpicture}%
}
\caption{CoSWA-YOLOv12. A compact YOLOv12 instance-segmentation network with a stride-four (P2) head (blue) gains three transfer-safe components: a wavelet detail residual in strided downsampling (green), a scale-adaptive NWD term in label assignment (orange), and a min-max Gaussian regression loss (purple). The inset curve shows the size gate opening for \pf and closing for larger classes, routing the Wasserstein treatment almost entirely to the ring-stage class. Each component reduces to the baseline at initialisation ($\gamma\!=\!0$, $\rho\!=\!0$, $\lambda\!=\!0$)}
\label{fig:arch}
\end{figure}

CoSWA-YOLOv12 is built on a YOLOv12 instance-segmentation network with an added stride-four (P2) detection head, a configuration previously shown to aid tiny parasite detection~\cite{anon_prior}. Figure~\ref{fig:arch} shows where the three additions attach: a wavelet detail residual inside the strided downsampling of the backbone, a scale-adaptive Wasserstein term inside the label assigner, and a min-max Gaussian regression loss at the box head. Each is engineered so that, at initialisation, it reproduces the baseline exactly, the property we exploit for lossless weight transfer.

\subsection{Wavelet Detail Residual Downsampling}
\label{sec:wdr}

Strided convolutions reduce spatial resolution by discarding high-frequency content, which is precisely the content that separates a small ring from background texture. We replace each strided downsampling convolution by its sum with a fixed Haar wavelet high-frequency branch,
\begin{equation}
\mathrm{WDR}(x) = \mathrm{Conv}_{s}(x) + \gamma \cdot \mathrm{Proj}\big(H(x)\big),
\end{equation}
where $\mathrm{Conv}_{s}$ is the original strided convolution whose weights carry over from the pretrained model, $H(\cdot)$ extracts the three Haar detail sub-bands, $\mathrm{Proj}$ is a $1{\times}1$ projection to the output channels, and $\gamma$ is a single learnable scalar. We initialise $\gamma=0$, so at the start of training $\mathrm{WDR}(x)=\mathrm{Conv}_{s}(x)$; optimisation then opens the detail branch only where, and as much as, it helps.

\subsection{Scale-Adaptive Wasserstein Assignment}
\label{sec:coswa}

\subsubsection{Wasserstein overlap.}
Following NWD~\cite{wang_normalized_2021}, a box $b=(c_x,c_y,w,h)$ is modelled as a $2$-D Gaussian, and the similarity between a ground-truth box and a candidate is
\begin{equation}
\mathrm{NWD}(b_1,b_2)=\exp\!\Big(-\tfrac{1}{C}\sqrt{W_2^2(b_1,b_2)}\Big),
\end{equation}
where $W_2^2$ is the squared $2$-Wasserstein distance between the two Gaussians, computed on the stacked centre and half-extent vector $(c_x,c_y,\tfrac{w}{2},\tfrac{h}{2})$, and $C$ is a dataset scale constant. Unlike IoU, $\mathrm{NWD}$ stays informative when boxes barely overlap, the regime that starves tiny objects during assignment.

\subsubsection{Size gate.}
The task-aligned assigner scores each candidate through an overlap term. We replace that term with a blend of CIoU and $\mathrm{NWD}$ whose mixing weight depends on the size of the ground-truth object,
\begin{align}
\mathrm{ov}(b_{gt},b) &= \big(1-r(b_{gt})\big)\,\mathrm{CIoU}(b_{gt},b) + r(b_{gt})\,\mathrm{NWD}(b_{gt},b),\\
r(b_{gt}) &= \rho\;\sigma\!\Big(\tfrac{\tau-\sqrt{w_{gt}\,h_{gt}}}{\beta}\Big),
\label{eq:gate}
\end{align}
where $\sqrt{w_{gt}h_{gt}}$ is the object size in pixels, $\sigma$ is the logistic function, $\rho\in[0,1]$ caps the maximum NWD weight, and $\tau,\beta$ set the midpoint and softness of the transition. Small objects ($\sqrt{wh}\ll\tau$) receive close to the full NWD weight $\rho$; large objects fall back to pure CIoU, and $\rho=0$ recovers the baseline everywhere. This single scalar per object is what lets one detector treat tiny objects without penalising larger ones on the same slide.

\subsection{Min-Max Gaussian Regression Loss}
\label{sec:m2nwd}

The regression loss must also stay informative at small scales. M2IoU sharpens IoU regression by penalising the box corner farther from its target more heavily through a coefficient $\alpha<0.5$~\cite{shandilya_m2iou_2024}; we carry this into the Gaussian domain. Writing $\mathrm{NWD}_{\min}$ and $\mathrm{NWD}_{\max}$ for the Wasserstein similarities on the nearer and farther corner, the $\text{M}^2$-NWD similarity and the box loss are
\begin{align}
\mathrm{M^2NWD} &= \alpha\,\mathrm{NWD}_{\min} + (1-\alpha)\,\mathrm{NWD}_{\max},\qquad \alpha < 0.5,\\
\mathcal{L}_{\text{box}} &= (1-\lambda)\big(1-\mathrm{CIoU}\big) + \lambda\big(1-\mathrm{M^2NWD}\big).
\end{align}
Because the Wasserstein similarity is scale-invariant, this term needs no size gate and is applied globally; $\lambda=0$ recovers pure CIoU regression.

\subsection{Transfer-Safety and Efficiency}
\label{sec:transfersafe}

Each component reduces to the baseline at initialisation ($\gamma=0$, $\rho=0$, $\lambda=0$), so loading pretrained weights yields a network numerically identical to the baseline, and no added machinery can degrade accuracy before it has learned to help, removing the usual risk of architectural change on a small dataset. The additions grow the model by only about one percent (Section~\ref{sec:setup}).

\subsection{Data, Compute, and Inference}
\label{sec:setup}

\subsubsection{Dataset and split.}
We evaluate on a five-class Rwandan thick-smear dataset of Giemsa-stained fields captured at $100\times$ magnification, annotated with boxes and polygon masks for four \textit{Plasmodium} species (\pf, \pmm, \po, \pv) and white blood cells (WBC)~\cite{anon_prior}. Images are letterboxed to $2048\times2048$ and split image-wise into $1{,}915$ training, $410$ validation, and $411$ test images ($\approx\!70/15/15$); the validation split is used only for checkpoint selection and all reported numbers are on the held-out test set. The classes span a wide size range, the median object side length being $34$ pixels for \pf, $78$ for \pmm, and $100$ to $115$ for \po, \pv, and WBC, the four-fold spread that makes a scale-uniform treatment counterproductive.

\subsubsection{Training and compute.}
All models are YOLOv12n instance-segmentation networks with a P2 head, trained for $70$ epochs at $2048\times2048$ with SGD (learning rate $0.005$, momentum $0.937$, weight decay $5\times10^{-4}$), mixed precision, batch size $4$, and mosaic and copy-paste augmentation~\cite{ghiasi_simple_2021}, with backbone and neck initialised from a public YOLOv12n segmentation checkpoint. We set the gate from our measured object sizes ($\rho=0.5$, $\tau=56$\,px, $\beta=8$\,px), and the loss coefficients are $\lambda=0.5$ and $\alpha=0.25$ (the M2IoU value~\cite{shandilya_m2iou_2024}); $\tau,\beta$ are training hyperparameters, set from the target dataset's object sizes. Training uses a single NVIDIA H100 GPU, the full model completing $70$ epochs in about $11.4$ hours. To attribute effects cleanly we compare seven configurations along a controlled factorial: the \textbf{baseline}; an \textbf{anti-aliased} downsampling control that smooths rather than preserves high frequencies; \textbf{WDR} alone; WDR with \textbf{global} NWD assignment; WDR with \textbf{scale-adaptive} assignment; WDR with global assignment and the $\text{M}^2$-NWD loss; and the \textbf{full} CoSWA-YOLOv12.

\subsubsection{Model size and inference.}
CoSWA-YOLOv12 has about $2.8$M parameters, only one percent above the baseline, and every added component is training-time or zero-initialised, so it adds no inference cost over the baseline. At $2048\times2048$ inference is $17.3$\,ms per image ($\approx\!58$ FPS, $19.5$\,ms end-to-end) on an NVIDIA A100, and the model also runs CPU-only in a Docker container for GPU-less deployment, keeping it within the footprint expected of an edge-deployable detector.

\section{Results}
\label{sec:results}

\subsection{Overall Performance}
\label{sec:overall}

Table~\ref{tab:overall} reports every standard metric for detection (box) and segmentation (mask) across the model ladder. CoSWA-YOLOv12 is strongest on the metrics that matter most for screening, with the best recall and mAP@50 on both tasks (mask $0.850$/$0.888$, box $0.851$/$0.889$); it trails only on precision, where the more conservative WDR-plus-global-NWD variant edges ahead at the cost of recall; CoSWA-YOLOv12 nonetheless improves precision over the baseline while raising recall, the harder and more useful direction. The anti-aliased control is uniformly weakest, confirming that preserving high-frequency detail, not smoothing it, helps tiny parasites.

\begin{table}[t]
\centering
\caption{Overall (all-class) test performance across the model ladder, for detection (box) and segmentation (mask). Best per column in bold. CoSWA-YOLOv12 (Full) leads recall, mAP@50, and the strict mAP@50-95 on both tasks, and improves precision over the baseline}
\label{tab:overall}
\setlength{\tabcolsep}{2pt}
\scriptsize
\begin{tabular}{l cccc cccc}
\toprule
 & \multicolumn{4}{c}{Detection (box)} & \multicolumn{4}{c}{Segmentation (mask)} \\
\cmidrule(lr){2-5}\cmidrule(lr){6-9}
Configuration & P & R & mAP50 & mAP50-95 & P & R & mAP50 & mAP50-95 \\
\midrule
Baseline                 & 0.802 & 0.827 & 0.869 & 0.642 & 0.803 & 0.828 & 0.871 & 0.630 \\
Anti-aliased             & 0.798 & 0.810 & 0.846 & 0.590 & 0.807 & 0.806 & 0.847 & 0.572 \\
WDR                      & 0.817 & 0.809 & 0.860 & 0.654 & 0.816 & 0.811 & 0.861 & 0.637 \\
WDR + global NWD         & \textbf{0.842} & 0.819 & 0.869 & 0.605 & \textbf{0.844} & 0.820 & 0.873 & 0.603 \\
WDR + scale-adaptive     & 0.801 & 0.822 & 0.867 & 0.639 & 0.801 & 0.822 & 0.868 & 0.629 \\
WDR + global + M$^2$-NWD & 0.826 & 0.823 & 0.877 & 0.634 & 0.830 & 0.827 & 0.882 & 0.621 \\
\textbf{CoSWA (Full)}    & 0.834 & \textbf{0.851} & \textbf{0.889} & \textbf{0.682} & 0.833 & \textbf{0.850} & \textbf{0.888} & \textbf{0.674} \\
\bottomrule
\end{tabular}
\end{table}

\begin{table}[t]
\centering
\caption{Per-class and overall segmentation (mask) performance, baseline vs.\ CoSWA-YOLOv12. Recall and mAP@50 (left) govern whether a parasite is found; mAP@50-95 (right) governs how tightly. Bold marks an improved value; detection (box) values track these within one to two points (Table~\ref{tab:overall})}
\label{tab:perclass}
\setlength{\tabcolsep}{6pt}
\footnotesize
\begin{tabular}{l cc cc cc}
\toprule
 & \multicolumn{2}{c}{Recall} & \multicolumn{2}{c}{mAP@50} & \multicolumn{2}{c}{mAP@50-95} \\
\cmidrule(lr){2-3}\cmidrule(lr){4-5}\cmidrule(lr){6-7}
Class & Base & Full & Base & Full & Base & Full \\
\midrule
\pf  & 0.629 & \textbf{0.741} & 0.734 & \textbf{0.807} & 0.456 & \textbf{0.492} \\
\pmm & 0.835 & \textbf{0.890} & 0.885 & \textbf{0.920} & 0.617 & \textbf{0.680} \\
\po  & 0.829 & 0.823 & 0.891 & \textbf{0.894} & 0.685 & \textbf{0.730} \\
\pv  & 0.896 & 0.871 & 0.924 & 0.887 & 0.628 & \textbf{0.665} \\
WBC  & 0.949 & 0.925 & 0.923 & \textbf{0.933} & 0.766 & \textbf{0.802} \\
\midrule
All  & 0.828 & \textbf{0.850} & 0.871 & \textbf{0.888} & 0.630 & \textbf{0.674} \\
\bottomrule
\end{tabular}
\end{table}

\subsection{Detecting the Critical Class: Recall and mAP@50}
\label{sec:detection}

The clinically decisive result is on \pf, on the metrics that decide whether a parasite is found at all. Table~\ref{tab:perclass} breaks the segmentation results down by class. On \pf, recall rises from $0.63$ to $0.74$ and mAP@50 from $0.73$ to $0.81$, an eleven-point recall gain on the class where recall is the safety-critical quantity, and \pmm improves in parallel ($0.84\to0.89$ recall, $0.89\to0.92$ mAP@50). The already-saturated \po, \pv, and WBC hold their high detection scores and register their gains on strict localisation (Section~\ref{sec:localisation}) instead. Averaged over all classes, recall improves from $0.828$ to $0.850$ and mAP@50 from $0.871$ to $0.888$, so the \pf gain does not come at the slide's expense. The validation confusion matrix makes the mechanism concrete: correctly detected \pf rises from $0.61$ to $0.85$ and \pf missed to background falls from $0.38$ to $0.15$, from two in five to under one in six, while cross-species confusion stays negligible throughout ($\leq 0.02$), so the difficulty is detection against background, not species discrimination.

\subsection{Strict Localisation: mAP@50-95}
\label{sec:localisation}

Where recall and mAP@50 measure whether a parasite is found, mAP@50-95 measures how tightly, and every class improves (Table~\ref{tab:perclass}, right): CoSWA-YOLOv12 is best on all five and overall, for both mask ($0.630\to0.674$) and box ($0.642\to0.682$). The gains are broadest on the classes already saturated on detection, \po ($0.685\to0.730$), \pv ($0.628\to0.665$), and WBC ($0.766\to0.802$): the framework tightens localisation on the larger classes while recovering the missing rings, which scale-uniform NWD cannot.

\subsubsection{Synergy and necessity of the gate.}
The assignment and loss innovations form a $2\times2$ factorial over the WDR backbone (Table~\ref{tab:overall}, mask mAP@50-95): the gate alone adds $0.026$ and the min-max loss alone $0.018$, but together $0.071$, a synergy of $0.027$, because with the gate in place the loss no longer spends capacity repairing large objects that global NWD mis-assigned. The gate is essential: sized from the data ($\tau{=}56$, $\beta{=}8$\,px) it opens to $0.94$ for \pf but below $0.06$ for every larger class, so removing it lets global NWD collapse overall mask mAP@50-95 from $0.637$ to $0.603$, which the gate restores before the $\text{M}^2$-NWD loss lifts every class above the baseline.

\section{Discussion}
\label{sec:discussion}

The results support this paper's central claim: on a multi-species slide the obstacle to detecting tiny \pf is not model capacity but a scale conflict inside training, and resolving it selectively is what unlocks the gains. The scale gate routes the Wasserstein treatment almost entirely to \pf, which recovers the recall and mAP@50 it was missing while the larger classes keep, and even tighten, their localisation; the factorial shows the gate and the min-max loss to be synergistic because the gate frees the loss from repairing mis-assigned large objects. Because box and mask move together throughout, the improvement must lie upstream of either head, in assignment and feature preservation.

\subsubsection{Limitations and future work.}
Several residual \pf false positives, on inspection, fall on unlabelled parasite-like structures, so the reported \pf precision is likely a lower bound. The single-site evaluation leaves cross-site robustness, a known weakness for tiny parasites, as the natural next steps.

\section{Impact in Resource-Constrained Settings}
\label{sec:impact}

Malaria's burden falls overwhelmingly on low-resource endemic regions where trained microscopists are scarce and confirmatory testing is often unavailable, so an automated reader on modest hardware can extend expert-level screening to where it is needed most. CoSWA-YOLOv12 is built for that: it keeps the $2.8$M-parameter, edge-deployable size and zero added inference cost of a nano detector, and its transfer-safety lets it adapt to a new laboratory from small local data without degrading. Most importantly, the gains land where clinical safety is decided, recall on ring-stage \pf: cutting missed \pf from $38\%$ to $15\%$ directly reduces missed diagnoses in settings with no second reader. The method also reaches well beyond this setting: its three components are scale-agnostic and transfer-safe at any YOLOv12 size, so any tiny-object detection task can adopt it directly.

\begin{credits}
\subsubsection{\discintname}
The authors have no competing interests to declare that are relevant to the content of this article.
\end{credits}

\newpage
\bibliographystyle{splncs04}
\bibliography{refs}

\end{document}